\documentclass[10pt]{article} 

\usepackage[preprint]{rlj} 

\usepackage{amssymb}
\usepackage{dsfont}
\usepackage{mathtools}
\usepackage{graphicx}
\usepackage{subcaption}
\usepackage[space]{grffile}
\usepackage{url}
\usepackage{booktabs}
\usepackage{multirow}
\usepackage{colortbl}
\newcolumntype{C}[1]{>{\centering\arraybackslash}m{#1}}
\usepackage{diagbox}

\usepackage{amsthm}
\newtheorem{definition}{Definition}
\newtheorem{property}{Property}

\usepackage{framed}
\usepackage{wrapfig}

\title{BXRL: Behavior-Explainable\\Reinforcement Learning}

\setrunningtitle{BXRL: Behavior-Explainable Reinforcement Learning}

\author{Ram Rachum\textsuperscript{1,2}, Yotam Amitai\textsuperscript{3}, Yonatan Nakar\textsuperscript{4}, Reuth Mirsky\textsuperscript{2,$\dagger$}, Cameron Allen\textsuperscript{1,$\dagger$}}
\emails{ram.rachum@berkeley.edu, yotamitai@gmail.com, yonatannakar@google.com, reuth.mirsky@tufts.edu, camallen@berkeley.edu}
\affiliations{
$^{1}$University of California, Berkeley\\
$^{2}$Tufts University\\
$^{3}$Independent Researcher\\
$^{4}$Google Research
\par
$^\dagger$ Equal senior contribution
}

\contribution{
    Behavior-Explainable Reinforcement Learning (BXRL): a formulation that allows practitioners to specify any behavior measure $m(\pi) : \Pi \to \mathbb{R}$ and request an explanation for why the policy is behaving that way.
}{
    Existing XRL methods explain specific actions, trajectories, policies, objectives or the training process. None allow freely specifying which behavioral pattern to explain.
}

\contribution{
    Analysis of three existing methods (data attribution, Shapley-based feature importance, and counterfactual policy explanation), showing how each can be retargeted at behavior measures with minimal structural changes.
}{
    No existing XRL method targets user-specified behaviors. We show that the BXRL formulation is practical by dissecting three methods and identifying the exact substitution needed to replace their original target functions with $m(\pi)$.
}

\contribution{
    HighJax: a JAX implementation of the HighwayEnv driving environment that provides an interface for defining behaviors, which can then be measured and differentiated with respect to the model parameters.
}{
    HighwayEnv is a popular environment used for RL and XRL.
}


\summary{
A major challenge of Reinforcement Learning is that agents often learn undesired behaviors that seem to defy the reward structure they were given. Explainable Reinforcement Learning (XRL) methods can answer queries such as ``explain this specific action'', ``explain this specific trajectory'', and ``explain the entire policy''. However, XRL lacks a formal definition for \textit{behavior} as a pattern of actions across many episodes. We provide such a definition, and use it to enable a new query: \textit{``Explain this behavior''}. We present \textbf{Behavior-Explainable Reinforcement Learning (BXRL)}, a new problem formulation that treats behaviors as first-class objects. BXRL defines a \textit{behavior measure} as any function $m : \Pi \to \mathbb{R}$, allowing users to precisely express the pattern of actions that they find interesting and measure how strongly the policy exhibits it. We define \textit{contrastive behaviors} that reduce the question ``why does the agent prefer $a$ to $a'$?'' to ``why is $m(\pi)$ high?'' which can be explored with differentiation. We do not implement an explainability method; we instead analyze three existing methods and propose how they could be adapted to explain behavior. We present a port of the HighwayEnv driving environment to JAX, which provides an interface for defining, measuring, and differentiating behaviors with respect to the model parameters.
}

\begin{document}

\makeatletter\if@preprint\else\makeCover\fi\makeatother
\maketitle

\begin{abstract}
A major challenge of Reinforcement Learning is that agents often learn undesired behaviors that seem to defy the reward structure they were given. Explainable Reinforcement Learning (XRL) methods can answer queries such as ``explain this specific action'', ``explain this specific trajectory'', and ``explain the entire policy''. However, XRL lacks a formal definition for \textit{behavior} as a pattern of actions across many episodes. We provide such a definition, and use it to enable a new query: \textit{``Explain this behavior''}. 

We present \textbf{Behavior-Explainable Reinforcement Learning (BXRL)}, a new problem formulation that treats behaviors as first-class objects. BXRL defines a \textit{behavior measure} as any function $m : \Pi \to \mathbb{R}$, allowing users to precisely express the pattern of actions that they find interesting and measure how strongly the policy exhibits it. We define \textit{contrastive behaviors} that reduce the question ``why does the agent prefer $a$ to $a'$?'' to ``why is $m(\pi)$ high?'' which can be explored with differentiation. We do not implement an explainability method; we instead analyze three existing methods and propose how they could be adapted to explain behavior. We present a port of the HighwayEnv driving environment to JAX, which provides an interface for defining, measuring, and differentiating behaviors with respect to the model parameters.
\end{abstract}

\section{Introduction}
\label{sec:intro}

A chatbot optimized with human feedback declares love to a journalist and expresses a desire to spread misinformation \citep{roose2023sydney}. An Autonomous Driving (AD) system powering a fleet of robotaxis learns to brake unnecessarily hard near bicyclists, increasing the risk of rear-end collisions \citep{zoox2025march_recall}. A swarm of warehouse robots navigates a grid to pick groceries, but three of its members collide into each other, causing a massive fire \citep{bbc2021fire}.

We want to understand \textit{why} these failures happen. Do the agents not receive enough information about their environment? Are the failures a result of out-of-distribution generalization? Or are the agents' actions an inevitable consequence of the incentive structure we provided \citep{knox2023reward}?

We advocate for explaining the recurring patterns of undesirable behavior that lead to failures, rather than the failures themselves. In our last example, the warehouse robots were observed to ``play chicken'', approaching each other at high speed and diverting at the last moment \citep{cnn2021_ocado_play_chicken}. In other domains, behaviors will be described not in terms of velocity and acceleration, but of defection rate, portfolio concentration, word choice, and more. \textbf{Given unwanted agent behavior of any kind, we wish to define it, measure it, and find an explanation for it.}

This need is within the scope of Explainable Reinforcement Learning (XRL), which is a subdomain at the intersection of Explainable Artificial Intelligence (XAI) and Reinforcement Learning (RL) \citep{heuillet2021explainability}. XRL aims to make the decisions of RL agents more interpretable and understandable to humans \citep{doshivelez2017interpretable}. It is driven by the needs to diagnose failures \citep{amodei2016concrete}, maintain user trust \citep{ribeiro2016lime}, satisfy regulations \citep{goodman2017european}, and ensure alignment with societal goals \citep{arrieta2020explainable}.

\citet{freiesleben2023dear} warn about a pervasive problem in XAI: explanation methods are routinely proposed without clearly defining what it is they aim to explain. This makes it difficult to invoke multiple explanation methods on the same phenomenon for a comparative study. This problem also exists in XRL, specifically for explanations of behavior, as distinct from actions or policies. When XRL papers use the term ``behavior'', they offer no formal definition for it, or any way to pinpoint one behavior out of many. Some papers use the word ``behavior'' as a synonym for policy. \citet{atrey2020exploratory} provide a verbal definition: \textit{``behavior, or aggregate actions, over temporally extended sequences of frames''}. However, a quantitative definition is needed to (1) measure \textit{how strongly} a given policy exhibits a given behavior and (2) allow the user to point at a specific behavior and ask ``why is the agent behaving like \textit{this}?''

This gap in problem formulation extends beyond method papers; no survey paper or position paper poses the problem of explaining behavior, nor provides a definition for behavior. We address this gap in this paper. We find a simple and general definition for behavior in an adjacent research area and repurpose it for explainability. We use it to make the following contributions:

\begin{enumerate}
    \item We formalize the notion of \textit{behavior} in RL as any scalar function $m : \Pi \to \mathbb{R}$ and introduce \textbf{Behavior-Explainable Reinforcement Learning (BXRL)}, a problem formulation that treats behaviors as first-class explanation targets. We show how to articulate specific behaviors as measurable quantities that, when differentiable, enable gradient-based analysis. (Sec.~\ref{sec:properties-behavior}--\ref{sec:bxrl}).
    \item We analyze three existing methods---data attribution, SVERL-P, and COUNTERPOL---and show how each can be adapted to analyze behavior measures with minimal structural changes (Sec.~\ref{sec:adapting-methods}).
    \item To further facilitate BXRL research, we introduce \textbf{HighJax}, a port of the HighwayEnv driving environment to JAX. HighJax provides an interface for defining, measuring, and differentiating behaviors with respect to the model parameters. We use HighJax to define an example behavior that leads to vehicle collisions, and invite researchers to use this environment to conduct BXRL research. (Sec.~\ref{sec:highjax}).
\end{enumerate}

\section{Background}
\label{sec:background}

\paragraph{Explanation target and source} \citet{hempel1965aspects} draws the distinction between the \textit{explanandum} (the phenomenon to be explained) and the \textit{explanans} (the information that does the explaining). We use the terms \textit{explanation target} and \textit{explanation source}. The former is the problem domain and the latter is the solution domain. Keeping with \citet{freiesleben2023dear}, we practice great care in keeping them separate.

\paragraph{Contrastive explanation} While philosophers have long debated what constitutes an explanation \citep{hume1748enquiry,lewis1973causation,woodward2003making}, research in cognitive science and social psychology has highlighted a key finding: good explanations are \textit{contrastive} \citep{lipton1990contrastive,miller2019explanation}. People do not ask ``Why $P$?'' in isolation; they ask ``Why $P$ rather than $Q$?'' even when $Q$ is implicit. $P$ and $Q$ are respectively called \textit{fact and foil}. For example, if a person asks ``why did my barn catch on fire?'' we recognize that the answer ``because it had oxygen in it'', while correct, has no explanatory power. The implicit ``...rather than $Q$?'' is ``...while other buildings in the area did not catch on fire?'' The answer to a contrastive question follows Lipton's \textit{Difference Condition}: to explain why $P$ rather than $Q$, cite a cause $C$ that is present in $P$'s causal history but absent from $Q$'s. In our example, $C$ might be ``because someone threw a lit cigarette in a pile of hay''. Crucially, the foil determines which causes are explanatory: the same fact $P$ has different explanations depending on which foil $Q$ the questioner has in mind.

\paragraph{Behavior as disposition} We ground our usage of the term ``behavior'' in \citet{ryle1949concept}'s dispositional analysis. Ryle described behaviors as tendencies to take certain actions in certain situations: ``to be a smoker is just to be bound or likely to fill, light and draw on a pipe in such and such conditions''. He distinguished single-track dispositions from multi-track dispositions, where multi-track dispositions could manifest as several different types of actions. This dispositional view is echoed in the psychological habit literature, where habits are characterized by ``activation by recurring context cues'' \citep{wood2016psychology}.

\paragraph{Markov Decision Process (MDP)} An MDP is defined by $(\mathcal{S}, \mathcal{A}, T, R, \gamma)$: states $\mathcal{S}$, actions $\mathcal{A}$, transition function $T(s'|s,a)$, reward function $R(s,a)$, and discount factor $\gamma$. At each timestep $t$, the agent observes state $s_t \in \mathcal{S}$, selects action $a_t \sim \pi_\theta(a \mid s_t)$ according to its \textit{policy} $\pi_\theta : \mathcal{S} \times \mathcal{A} \to [0,1]$, a conditional distribution over actions given states, parameterized by $\theta$, receives reward $r_t$, and the environment transitions to a new state. A trajectory $\tau = (s_0, a_0, r_0, s_1, \ldots)$ is a sequence of experiences; an epoch is a batch of $N$ timesteps collected for a single gradient update. The return $G_t = \sum_{k=0}^{\infty} \gamma^k r_{t+k}$ is the discounted sum of future rewards. The agent's goal is to maximize expected return, or value $V(s) = \mathbb{E}[G_t | s_t = s]$. The advantage $A_t = G_t - V(s_t)$ measures how much better an action was than expected. The training objective $J(\pi) = \mathbb{E}[G_0]$ is the expected return under $\pi$. The Kullback--Leibler (KL) divergence $D_{\mathrm{KL}}(\pi_{\text{old}} \| \pi_{\text{new}}) = \mathbb{E}_{\pi_{\text{old}}}[\log \frac{\pi_{\text{old}}}{\pi_{\text{new}}}]$ is commonly used to constrain policy updates so that each gradient step does not change the policy too drastically \citep{kullback1951information}. We note that BXRL applies to Partially-Observable MDPs as well, where $o_t$, the \textit{observation} the agent receives at timestep $t$, is distinct from $s_t$; in a fully-observable MDP, $o_t = s_t$.

\paragraph{Attribution and importance methods} Several methods attribute a scalar objective to individual components of the learning process. \textit{Data attribution} \citep{hu2025snapshot} adapts TracIn \citep{pruthi2020tracin} to identify which training records most influenced a target function, measuring influence via the gradient inner product between the target and the PPO loss \citep{schulman2017proximal}. \textit{SVERL-P} \citep{beechey2023sverl} considers Shapley values over state features, quantifying each feature's contribution to agent performance. \textit{COUNTERPOL} \citep{deshmukh2023counterfactual} finds the minimal policy change that achieves a specified target return, optimizing a counterfactual objective with a KL-divergence proximity constraint. All three are built around a modular scalar objective that can be retargeted at behavior measures.

\section{Properties of Behavior}
\label{sec:properties-behavior}

We distill the single-track model of behavior presented by \citet{ryle1949concept} to concrete MDP properties:

\begin{framed}
\begin{property}[Action-based]
\label{prop:action-based}
Behavior consists of one or more actions the agent tends to choose.
\end{property}

\begin{property}[Recurring]
\label{prop:recurring}
Behavior recurs in multiple episodes.
\end{property}

\begin{property}[Graded]
\label{prop:graded}
A behavior is not merely present or absent; a policy may exhibit a behavior strongly, weakly, or anywhere in between.
\end{property}

\begin{property}[Scoped]
\label{prop:scoped}
Any given behavior will be limited in scope in some way: it may be scoped to a specific set of states, or to specific dimensions of the action space if it is multi-dimensional. Timesteps in states outside this scope are irrelevant to the behavior, as are action dimensions outside it.
\end{property}

\end{framed}

We previously used the motivating example of a warehouse robot that plays chicken, i.e., moves too fast toward human workers. This phenomenon satisfies all four properties: (1) it's expressed by the action of speeding up, (2) it happens in separate incidents, (3) it can happen less often or more often, with varying speeds, and (4) it may only occur in states in which the robot is moving toward humans, and only actions related to full-body movement are relevant to it.

In the next section we map out the XRL landscape and find a gap in its treatment of behavior.

\section{Related Work}
\label{sec:related-work}

We are interested in the subset of XRL literature that is guided by a strong motivation to answer a focused question, giving preference to ``why?'' questions. We deem as out of scope any XRL paradigm whose primary goal is to build trust, verify safety properties, or summarize policies. We organize the remaining methods by what kind of question they answer, i.e., their \textit{explanation target}.\footnote{Some papers explicitly define their explanation target or the questions that they answer, while some leave it implicit; there can be different interpretations on what is the primary question that a method answers.} We choose three surveys that provide a taxonomy based on explanation target \citep{vouros2023explainable,milani2024explainable,saulieres2025survey} and pool them into a cross-taxonomy of five distinct explanation targets, shown in Table~\ref{tab:targets}.

\begin{table}[b]
\centering
\definecolor{tablerule}{gray}{0.65}%
\definecolor{targetcol}{RGB}{255,252,225}%
\definecolor{surveycol}{RGB}{240,240,255}%
{\arrayrulecolor{tablerule}%
\setlength{\arrayrulewidth}{0.4pt}%
\renewcommand{\arraystretch}{1.5}%
\makebox[\textwidth][c]{%
\resizebox{\textwidth}{!}{
\begin{tabular}{|C{1.5cm}|>{\raggedright\arraybackslash}m{5.5cm}|C{1.8cm}|C{2.9cm}|C{1.5cm}|C{1.1cm}|>{\raggedright\arraybackslash}m{2.8cm}|}
\arrayrulecolor{black}\hline\arrayrulecolor{tablerule}
\rowcolor{black!8}
\footnotesize{\textbf{Explanation Target}} & \multicolumn{1}{C{5.5cm}|}{\textbf{Example questions}} & \textbf{\citet{vouros2023explainable}} & \textbf{\citet{milani2024explainable}} & \footnotesize{{\textbf{\citet{saulieres2025survey}}}} & \footnotesize{\textbf{\# papers surveyed}} & \multicolumn{1}{C{2.8cm}|}{\textbf{Example methods}} \\
\arrayrulecolor{black}\hline\arrayrulecolor{tablerule}
\cellcolor{targetcol}\textbf{Action} & ``In episode 137, timestep 82, why did the agent choose action $a$?'' & \cellcolor{surveycol}Outcome & \cellcolor{surveycol}{\scriptsize Feature Importance: Directly Generate Explanations} & \cellcolor{surveycol}Action & $105$ & \footnotesize{\citet{puri2019explain,juozapaitis2019reward}} \\
\hline
\cellcolor{targetcol}\textbf{Trajectory} & ``In episode 137, which timesteps were critical for the outcome?'' & \cellcolor{surveycol}--- & \cellcolor{surveycol}--- & \cellcolor{surveycol}Sequence & 11 & \footnotesize{\citet{tsirtsis2021counterfactual,sreedharan2022bridging}} \\
\hline
\cellcolor{targetcol}\textbf{Policy} & ``What is the agent's general decision-making process?'' & \cellcolor{surveycol}Policy & \cellcolor{surveycol}{\footnotesize FI: Interpretable Policies, Policy-Level} & \cellcolor{surveycol}Policy & $192$ & \footnotesize{\citet{verma2018programmatically,bastani2018verifiable}} \\
\hline
\cellcolor{targetcol}\textbf{Objective} & ``What is the agent optimizing? What are its goals?'' & \cellcolor{surveycol}Objectives & \cellcolor{surveycol}--- & \cellcolor{surveycol}--- & 3 & \footnotesize{\citet{huang2019enabling,bica2021learning}} \\
\hline
\cellcolor{targetcol}\textbf{Learning} & ``Why did the policy gradually evolve to be the way that it is?'' & \cellcolor{surveycol}--- & \cellcolor{surveycol}\footnotesize{Learning Process \& MDP} & \cellcolor{surveycol}--- & 17 & \footnotesize{\citet{dao2018deep,wang2018dqnviz}} \\
\hline
\end{tabular}%
}
}%
\caption{A cross-taxonomy of XRL methods according to their explanation targets. The purple cells indicate the corresponding categories in three surveys.}
\label{tab:targets}
\arrayrulecolor{black}}%
\end{table}

We seek an explanation target that aligns with our motivation of explaining behavior, as we described in Section~\ref{sec:properties-behavior}. \textbf{We argue that none of the five explanation targets in Table~\ref{tab:targets} are a good fit for explaining behavior:}

\begin{itemize}
    \item The \textit{objective} and \textit{learning} explanation targets violate Property~\ref{prop:action-based}; they are not directed at the agent's actions. While actions are involved, \textit{explaining} the actions is not their primary motivation.
    \item The \textit{action} and \textit{trajectory} explanation targets are targeted at a single timestep and a single episode respectively; this means that their respective methods could be useful for explaining a \textit{case study} of the chicken behavior, but that explanation might overfit the circumstances of the specific episode, and therefore violates Property~\ref{prop:recurring}. In that sense, these explanation targets are too narrow.
    \item The \textit{policy} explanation target is too broad; the policy is responsible for all the activities that the agent does in all states. Our warehouse robot may do a great job at stacking boxes and scanning barcodes, but given Property~\ref{prop:scoped}, these action components are out of scope, and any breakdown of them is noise that we do not want in our explanation.
\end{itemize}

Property~\ref{prop:recurring} is especially important in MDPs that favor stochastic policies. Consider a music recommender system that has access to millions of songs. It performs well, except that for 10\% of user queries it plays \textit{The Final Countdown} by Europe. This is a bug that the operators would urgently want explained. However, the \textit{action} explanation target would not be useful, not because the methods are not good enough, but because on the action level, \textit{there is nothing to explain}. The song itself is fine. It is only the aggregate behavior of playing it often that calls for an explanation.

\textbf{None of the surveyed XRL methods allow users to receive an explanation for a specified behavior;} and to the best of our knowledge, no XRL survey defines a general expression for behavior to be explained \citep{amitai2024survey}. We note three relevant methods below. None of them allow users to specify a behavior, measure it, or answer ``why did this behavior occur?'' but each addresses a related aspect of behavior analysis in RL.

\paragraph{\citet{rishav2025behavior}} answer the question ``what behavior is this action part of?'' The authors use behaviors as an explanation \textit{source}, while their explanation target is actions. They do not allow users to choose a specific behavior; they instead (1) detect behaviors automatically by using a VQ-VAE to discretize offline state-action trajectories into latent codes and (2) apply spectral clustering to discover recurring behavioral patterns. They then explain actions by attributing them to the discovered behaviors via per-cluster behavior cloning models.

\paragraph {HIGHLIGHTS \citep{amir2018highlights}} answers the question ``what timesteps give the most information about the policy's capabilities and limitations, and what actions does the policy choose in them?'' It summarizes an agent's policy by extracting trajectories through states where different actions lead to substantially different outcomes, as measured by Q-value gaps.

\paragraph{ASQ-IT \citep{amitai2024asqit}} answers the question ``in scenarios that match the constraints I defined, what actions does the agent choose?'' They provide an interface for users to specify start/end conditions and constraints on what happens in between. Though ASQ-IT provides explanations through demonstration, it does not define or measure behaviors as general constructs, and its query language uses discrete boolean predicates, so gradual or continuous behavioral properties are not expressible. They demonstrate on a highway domain where the user can specify, e.g., that the agent starts in Lane~1 behind another car and ends in Lane~4. ASQ-IT then searches pre-collected rollouts and plays back matching trajectories.

The authors of the latter work call for ``a language for describing more abstract queries about the behavior of the agent''. We present such a formalism in the next two sections.

\section{Behavior-Explainable Reinforcement Learning (BXRL)}
\label{sec:bxrl}

We find a simple and general definition of behavior in the research area of evolutionary computation. \citet{lehman2011abandoning} introduced \textit{behavioral characterization}: a domain-dependent function that maps policies to a vector space, enabling search for behaviorally diverse solutions. \citet{pugh2016quality} formalized this as practitioner-selected features defining ``a space of possible behaviors'', and \citet{meyerson2016learning} made it an explicit, first-class learnable component. \citet{conti2018improving} applied this formalism to deep RL. \citet{knox2023reward} proposed sanity checks for reward functions that rely on policy-level performance metrics, which can be viewed as behavior measures. Adapting the notation of \citet{fontaine2021differentiable}, we denote behavior measures by $m(\pi)$ and define them as follows:

\begin{framed}
\begin{definition}[Behavior measure]
\label{def:behavior-measure}
Given an MDP, a behavior measure is any function $m$ from a policy $\pi$ to a scalar:
\[
m : \Pi \to \mathbb{R}
\]
A larger value of $m(\pi)$ indicates that $\pi$ exhibits the represented behavior more strongly.
\end{definition}
\end{framed}

A policy is a conditional probability distribution over actions given observations; this means any $m$ can be defined only in terms of how likely an agent is to take each of the actions, and not any other information. This definition satisfies the four properties we distilled from \citet{ryle1949concept} in Section~\ref{sec:properties-behavior}: (1) it is defined in terms of action probabilities, (2) it is not limited to a specific episode or timestep, (3) it returns a scalar that expresses how strongly the behavior is exhibited, and (4) it allows defining behaviors that are dependent on only a subset of the MDP's observation space and action space. We show practical examples of such behaviors in Section~\ref{sec:express-behavior}.

We wish to highlight an important feature of Definition~\ref{def:behavior-measure}. We motivated our work with examples of real-world negative consequences of an agent's actions, such as traffic accidents. While these negative events may be modeled as trajectories in the MDP, our definition of behavior measure does not depend on any consequence or event, positive or negative, that the agent's actions may cause. It is defined strictly in terms of the policy, i.e. the probability that the agent will take an action that we associate with the bad event. It does not require sampling the policy or the state transition function. This has three desirable effects: (1) it does not require running an expensive rollout per-evaluation, (2) it does not add another source of variance, and (3) it allows defining behaviors that are fully differentiable with respect to the agent's parameters. The latter point is required for the adaptation of data attribution methods to explainability, detailed in Section~\ref{sec:adapt-data-attribution}.

One way to understand behavior measures is to view the training objective $J(\pi) \to \mathbb{R}$ as a measure over policies. $J$ is determined by the reward function $R(s,a)$. Practitioners design $R$ to produce a $J$ such that $\pi^*_J$ behaves in particular desirable ways, \textit{without having to specify each of these ways.} For example, we reward a bipedal robot for forward velocity, and it learns to maintain upright balance and a walking gait without us having to formalize these behaviors. We advocate for formalizing these behaviors, e.g., upright balance as $m_b(\pi)$ and walking gait as $m_g(\pi)$, and then using them as explanation targets, while keeping optimization on $J$ only. This allows substituting the question ``why is my robot not maintaining an upright balance?'' with the equivalent ``why is $m_b(\pi)$ low?'' The latter question allows for differentiation of $m_b$ with respect to policy parameters, and by extension, to any quantities that affect the policy parameters during training. We present a general definition for any method that attempts to answer questions of this form:

\begin{framed}
\begin{definition}[BXRL Method]
\label{def:bxrl-method}
Given a MDP and a policy $\pi$, a BXRL method is one that produces an explanation for any specified behavior measure $m$.
\end{definition}
\end{framed}

We intentionally leave the ``explanation'' output undefined. We wish to fix the explanation target, not its sources or output modalities. A few possible output formats are saliency maps, tables, plots, and text. We suggest that the way to compare methods that provide different types of explanations is by evaluating them in user studies that produce objective success metrics \citep{gyevnar2025objective}.

\section{Expressing a Behavior as a Number}
\label{sec:express-behavior}

Here we provide practical guidance on defining $m(\pi)$, with concrete examples in several MDP domains. In all our examples below, we follow the convention that $m(\pi) = 0$ indicates the behavior does not occur at all, and $m(\pi) = 1$ indicates the policy exhibits the behavior as strongly as possible. This convention is optional. We note that in simple domains, it is easy to find canonical metrics for behaviors; in complex domains, many correlating measures may exist for the same core behavior. We demonstrate this for several domains in Table \ref{tab:m-examples}.

\begin{table}[h]
\caption{Examples of expressing behaviors in different domains as a measure $m(\pi)$.}
\label{tab:m-examples}
\centering
\makebox[\textwidth][c]{%
\begin{tabular}{p{3.7cm}p{2.2cm}p{6.9cm}}
\toprule
\textbf{Domain} & \textbf{Behavior} & \textbf{Metric} \\
\midrule
\small{Iterated Prisoner's Dilemma} & Reciprocity & $m(\pi) := \mathbb{E}_{o \sim \mathcal{D}}[\pi_\theta(a_{t-1}^{\text{opp}} \mid o)]$ \\
Autonomous vehicles & Tailgating & $m(\pi) := \mathbb{E}_{o \sim \mathcal{D}_{\text{close}}}[1 - \pi_\theta(\textsc{slower} \mid o)]$ \\
LLMs & Sycophancy & $m(\pi) := \mathbb{E}_{x \sim \mathcal{D}}[\pi_\theta(\text{\footnotesize{``You're right to push back''}} \mid x)]$ \\
Robotics & Jerkiness & $m(\pi) := \mathbb{E}_{(o,o') \sim \mathcal{D}_{\text{consec}}}[\|\mu_\theta(o') - \mu_\theta(o)\|^2]$ \\
Trading & Risk appetite & $m(\pi) := \mathbb{E}_{o \sim \mathcal{D}}[|\mu_\theta(o)|]$ \\
Recommenders & Popularity bias & $m(\pi) := \mathbb{E}_{o \sim \mathcal{D}}[\sum_{i \in \text{top-}k} \pi_\theta(i \mid o)]$ \\
\bottomrule
\end{tabular}%
}
\end{table}

All metrics in Table~\ref{tab:m-examples} share a common structure: each is an expectation over a fixed observation distribution $\mathcal{D}$ of some function of the policy's output. We write $\pi_\theta(a \mid o)$ for the action probabilities of a discrete policy and $\mu_\theta(o)$ for the mean action of a continuous policy. Crucially, $\mathcal{D}$ is not dependent on $\theta$: it is a pre-collected or manually constructed set of observations. This ensures that the full computation from $\theta$ to $m(\pi_\theta)$ passes only through the policy network, making it end-to-end differentiable. When the behavior is scoped to specific situations (Property~\ref{prop:scoped}), this is reflected by restricting $\mathcal{D}$ accordingly; e.g., for a highway environment, $\mathcal{D}_{\text{close}}$ contains only observations where the ego vehicle is near the vehicle ahead. Practitioners may draw observations from rollouts of a reference policy, from a replay buffer, or construct them synthetically.

\citet{miller2019explanation} advocates for explanations that are contrastive, meaning that they answer questions of the form ``why P and not Q?'' We show how to define \textit{contrastive behavior measures} of two kinds in Table \ref{tab:contrastive-kinds}. \textit{Observation-contrasting behavior measures} are similar to classic fact-foil examples such as ``why did my barn catch on fire, while my neighbor's barn did not?'' \textit{Action-contrasting behavior measures} express questions such as ``why did the driving agent choose to overtake rather than stay in its lane?'' Both types allow reducing such questions to ``why does $m(\pi)$ have a high value?''

\begin{table}[h]
\caption{Two kinds of contrastive behavior measures.}
\label{tab:contrastive-kinds}
\centering
\makebox[\textwidth][c]{%
\begin{tabular}{p{1.9cm}p{6.8cm}p{4.1cm}}
\toprule
\textbf{Contrast} & \textbf{Question} & \textbf{Metric} \\
\midrule
Observation & \small{``How much more likely is the agent to choose action $a$ given observation $o_P$ than given observation $o_Q$?''} & \small{$m(\pi) := \pi(a \mid o_P) - \pi(a \mid o_Q)$} \\
Action & \small{``Given observation $o$, how much more likely is the agent to choose action $a_P$ rather than $a_Q$?''} & \small{$m(\pi) := \pi(a_P \mid o) - \pi(a_Q \mid o)$} \\
\bottomrule
\end{tabular}%
}
\end{table}

\section{Adapting Existing XRL Methods for BXRL}
\label{sec:adapting-methods}

Some existing methods can be retargeted at behavior measures, even if they were not originally designed for that purpose. We review one attribution method, one importance method, and one contrastive method. All three are built around a modular scalar objective that can be replaced with $m(\pi)$ with minimal structural changes. We do not implement these adaptations; we analyze each method's structure to show that the substitution is feasible. We analyze the changes that occur when $m(\pi)$ is substituted in, what properties are preserved, and what breaks.

\subsection{Data Attribution}
\label{sec:adapt-data-attribution}

\citet{hu2025snapshot} introduced a data attribution framework for online RL that answers the questions "which training experiences most influenced the agent's tendency to take action $a$ in state $s$?" and "which training experiences most increased the agent's overall performance?" The authors adapt TracIn \citep{pruthi2020tracin} to operate within single PPO training rounds. The method has two target functions. The first, $f^{\text{action}}(\theta) = \log \pi_\theta(a \mid s)$, targets the log-probability of a specific action in a specific state; influence scores quantify how much each training record pushed the policy toward or away from that action. The second, $f^{\text{return}}(\theta)$, is a REINFORCE-like estimate of expected return; influence scores quantify each record's contribution to overall performance. In both cases, the components receiving attribution are individual training records in the current round's rollout buffer. Within each round $k$, the influence of a record $z_i = (s_i, a_i, r_i, \log \pi_i, v_i, \hat{A}_i)$ on the target is measured via gradient similarity:
\begin{equation}
I_i = \left\langle \nabla_\theta f(\theta^{(k)}), \; \nabla_\theta \mathcal{L}^{\text{PPO}}(\theta^{(k)}, z_i) \right\rangle,
\end{equation}
where $f \in \{f^\text{action}, f^\text{return}\}$ and the other side of the dot product is always the per-record PPO training gradient.

The method's two targets serve different purposes. The action target quantifies which records most influenced a specific decision. The return target identifies training records that negatively influence expected return (e.g., via inaccurate advantage estimates); filtering out these records can help speed up learning. The action target is close to describing behavior in the general way that we define it; we suggest generalizing beyond a single state-action pair to (1) prevent overfitting to a single state and (2) support multi-state behaviors such as overtaking. (Expanded in Section~\ref{sec:limitations}.)

\paragraph{BXRL adaptation} Replace the target function $f(\theta)$ with any differentiable behavior measure $m(\pi_\theta)$, such that the influence score becomes $I_i = \langle \nabla_\theta m(\pi_\theta), \nabla_\theta \mathcal{L}^{\text{PPO}}(\theta, z_i) \rangle$. The influence scores then quantify which training transitions in round $k$ most contributed to increasing or decreasing $m(\pi)$. Positive scores indicate records that increased $m(\pi)$; negative scores indicate records that suppressed it. For behavior measures of the form $m(\pi) = \mathbb{E}_{o \sim \mathcal{D}}[g(\pi_\theta(\cdot \mid o))]$, where $g$ is a differentiable function, the gradient can be straightforwardly computed via backpropagation through $g \circ \pi_\theta$. The adaptation produces per-record influence scores that can be aggregated temporally to reveal \textit{when} during training a behavior emerged, and analyzed semantically to reveal \textit{what those experiences have in common}.

\paragraph{Limitations} This method reveals a record's contribution to only the \textit{current} round's policy update.  However, a policy update in one round could set the agent up to collect better data in subsequent rounds, and the method won't measure these downstream effects. \citet{hu2025snapshot} acknowledge that the local framework ``lacks a clear counterfactual interpretation''; rather it tells only whether a record's gradient pointed in a helpful/harmful direction. The underlying TracIn formulation is derived SGD, but PPO implementations use Adam.

\subsection{Shapley Values for Observation Features}
\label{sec:adapt-sverl}

SVERL-P \citep{beechey2023sverl} answers the question ``which observation features does the agent rely on most for its performance?'' It uses Shapley values to decompose RL agent performance into per-feature contributions. Its target quantity is expected return, and the components receiving credit are state features. The method works by comparing the agent's performance under a variety of modified policies, where each modification restricts the set of state features that the policy is allowed to condition on (marginalizing over the rest). For each feature $i$ and coalition of remaining features $C\subseteq F \setminus \{i\}$, it compares the agent's marginalized policy $\pi_C(a |s) := \sum_{s'\in S} p^\pi(s'|s_C) \pi(a|s')$, that ``ignores'' all the other features, against the marginalized policy $\pi_{C\cup\{i\}}$ that also conditions on feature $i$. It then uses the marginal gain in expected return to compute Shapley values $\phi_i$, assigning each feature the unique fair credit allocation that satisfies all four Shapley axioms: efficiency, symmetry, null player, and additivity.

The method has two versions: \textit{global} and \textit{local}. The \textit{global} version modifies the policy at all states and computes the expected contribution of each feature under the state occupancy distribution $p^\pi(s)$; the \textit{local} version only modifies the policy at a given state $s$, and computes expected return \textit{from state} $s$. Together these quantify how much each state feature contributes to the agent's performance. Note that SVERL-P measures the informational value of only the features for making good decisions, and can thus assign zero importance to a feature that's deeply embedded in the dynamics if the feature isn't decision-relevant for maximizing expected return.

\paragraph{BXRL adaptation} Replace the expected return with a behavior measure $m(\pi)$. The target quantity becomes the behavior measure, and the Shapley decomposition quantifies how much observing each feature $i$ contributes to $m(\pi)$. This adaptation produces scores such as ``the distance-to-vehicle-ahead feature contributes $+0.23$ to $m(\pi)$, while relative speed contributes $-0.11$.'' It extends SVERL-P to account for other behaviors, not just those that are relevant to expected-return, while retaining the theoretical guarantee of fair credit allocation across features.

\paragraph{Limitations} Exact computation is $O(2^{|F|} \cdot |S|)$, limiting applicability to environments with small feature spaces ($\leq 20$ features or so) without approximation. The original paper validates only on tabular MDPs with exact solutions, not neural network policies. FastSVERL \citep{beechey2025fastsverl} addresses scalability by amortizing costs across states with a parametric model, but this introduces approximation error.

\subsection{Counterfactual Policy Explanation}
\label{sec:adapt-counterpol}

COUNTERPOL \citep{deshmukh2023counterfactual} answers the question ``what is the smallest change to this policy that would make it perform at level $R_{\text{target}}$?'' This can be helpful for understanding which actions in which states provide the biggest opportunity for increasing or decreasing performance. Given a policy $\pi_{\theta_0}$ with expected return $J_{\pi_{\theta_0}}$ and a target return $R_{\text{target}}$, the method optimizes:
\begin{equation}
\pi_{\theta_{\text{cf}}} = \arg\min_\theta \left[ |J_{\pi_\theta} - R_{\text{target}}| + k \cdot D_{\mathrm{KL}}(\pi_{\theta_0} \| \pi_\theta) \right],
\end{equation}
where the first term pushes toward the target return and the KL-divergence term keeps the counterfactual policy close to the original. The coefficient $k$ regulates the trade-off. The return $J_{\pi_\theta}$ and its gradient are estimated via on-policy Monte Carlo rollouts and the policy gradient theorem. To reach distant targets, the method iteratively updates the KL pivot after every $m$ gradient steps, allowing the counterfactual to ``walk'' away from the original in controlled increments.

The output is a counterfactual policy $\pi_{\theta_{\text{cf}}}$ that achieves approximately $R_{\text{target}}$ while remaining as similar as possible to $\pi_{\theta_0}$. The structured difference $\pi_{\theta_{\text{cf}}}(\cdot \mid o) - \pi_{\theta_0}(\cdot \mid o)$ across observations reveals which states required different action distributions and how they changed. The authors show a theoretical connection to trust-region policy optimization: when $R_{\text{target}}$ equals the maximum possible return, the COUNTERPOL objective reduces to the TRPO objective.

\paragraph{BXRL adaptation} Replace the expected return $J_{\pi_\theta}$ with a behavior measure $m(\pi_\theta)$, and replace the target return $R_{\text{target}}$ with a target behavior level $m^*$. The counterfactual objective becomes:
\begin{equation}
\pi_{\theta_{\text{cf}}} = \arg\min_\theta \left[ |m(\pi_\theta) - m^*| + k \cdot D_{\mathrm{KL}}(\pi_{\theta_0} \| \pi_\theta) \right].
\end{equation}
Because behavior measures are pure functions of $\pi$, no rollouts are needed, making the adapted version simpler than the original. The counterfactual policy reveals what minimal policy change would increase or decrease the behavior: for example, ``to reduce the collision behavior $m_c$ from $0.7$ to $0.1$, the policy needs to shift probability from \textsc{faster} to \textsc{slower} in states where the nearest vehicle is close ahead, while other states remain essentially unchanged.''

\paragraph{Limitations} Unlike the previous two methods, COUNTERPOL does not decompose $m(\pi)$ into per-component contributions; it produces a modified policy whose difference from the original must be inspected to extract insight. The optimization may converge to one of several equally-minimal counterfactual policies, and different initializations may yield different solutions. The KL coefficient $k$ requires tuning per domain.

\section{Demonstrating Behaviors in Autonomous Driving}
\label{sec:highjax}

We demonstrate defining and measuring behaviors in a driving environment. HighwayEnv \citep{leurent2018highway} is a widely used driving simulation for RL research, but its NumPy implementation cannot be differentiated through. We introduce \textbf{HighJax}, a port of HighwayEnv's \texttt{highway-v0} setting to JAX \citep{jax2018github}. The HighJax state transition function $T(s'|s,a)$ is implemented as a sequence of pure matrix operations. When using HighJax, rollout and training are batched and just-in-time compiled for high efficiency and throughput. We provide HighJax under an MIT license at \url{https://github.com/HumanCompatibleAI/HighJax}.

\begin{wrapfigure}{r}{0.4\textwidth}
\centering
\vspace{-1em}
\includegraphics[width=0.38\textwidth]{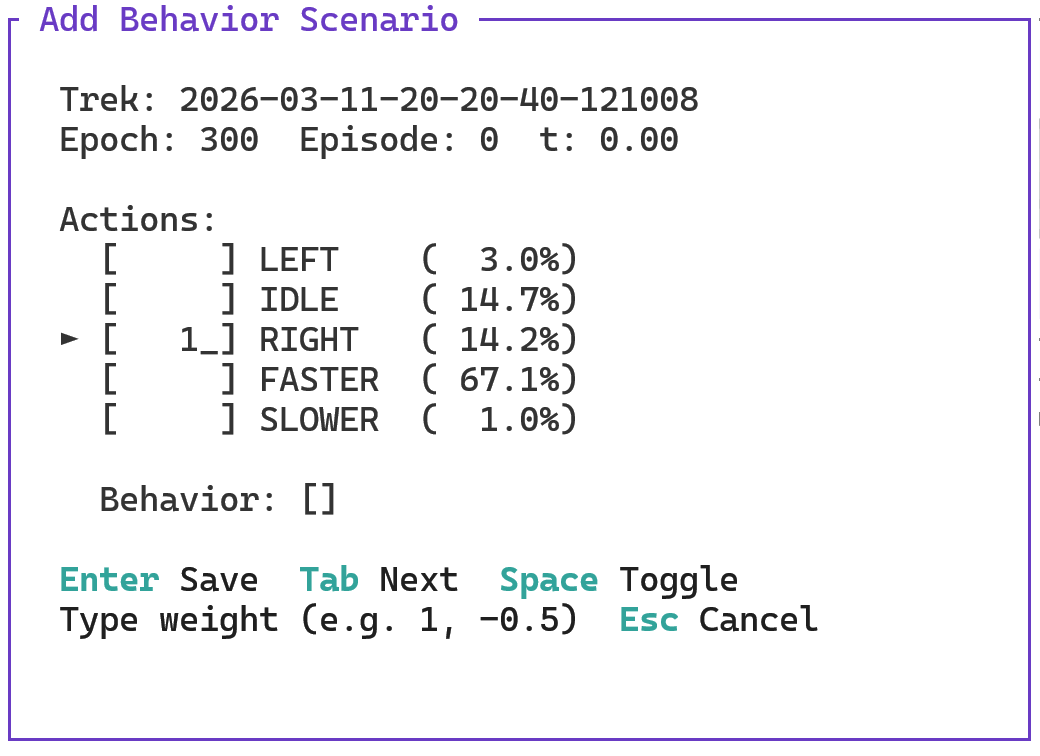}
\caption{HighJax TUI for defining behavior scenarios by selecting states and actions from existing rollouts.}
\label{fig:tui}
\vspace{-1em}
\end{wrapfigure}

HighJax provides a terminal user interface (TUI) that allows defining a behavior by selecting states and actions from existing rollouts (Fig.~\ref{fig:tui}). We define a behavior measure $m_c$ by choosing states in which the car veers towards other cars. We show how $m_c$ changes during training, and present it as an example target for BXRL methods.

\subsection{Environment rules}
\label{sec:highjax-env}

HighJax is a port of the \texttt{highway-v0} setting in which the agent controls a vehicle navigating a four-lane highway with fifty other vehicles ahead. The agent's vehicle is also called the \textit{ego vehicle}.

The \texttt{highway-v0} setting has a discrete action space: \textsc{left}, \textsc{idle}, \textsc{right}, \textsc{faster} and \textsc{slower}. These are interpreted by the environment into smooth movements. The other vehicles are controlled by three predefined algorithms: Intelligent Driver Model (IDM) \citep{treiber2000idm}, MOBIL \citep{kesting2007mobil}, and a proportional controller. The agent is rewarded for avoiding collisions, driving fast, and staying as close to the rightmost lane as possible. The reward at each timestep is
\begin{equation*}
    r_t = -1 \cdot \mathds{1}[\text{collision}] + 0.4 \cdot \tilde{v}_t + 0.1 \cdot \tilde{\ell}_t
\end{equation*}
where $\tilde{v}_t = \text{clip}((v_t - 20)/10,\; 0,\; 1)$ is the normalized forward speed (0 at $\leq$20\,m/s, 1 at $\geq$30\,m/s) and $\tilde{\ell}_t = \ell_t / 3$ is the normalized lane index (0 for leftmost, 1 for rightmost). For benchmarking we report the normalized reward $(r_t + 1)/1.5 \in [0, 1]$. The observation is a matrix $\mathbf{O} \in \mathbb{R}^{5 \times 5}$ with the following data for the ego vehicle and the four closest vehicles ahead: presence, $x$, $y$, $v_x$, and $v_y$. The rows for non-ego vehicles are in relative coordinates. Full details of the environment rules are given in Appendix~\ref{app:env-details}. 

\subsection{Analyzing Training Using Behavior}
\label{sec:analyzing-training}

\begin{figure}[b]
\centering
\includegraphics[width=\textwidth]{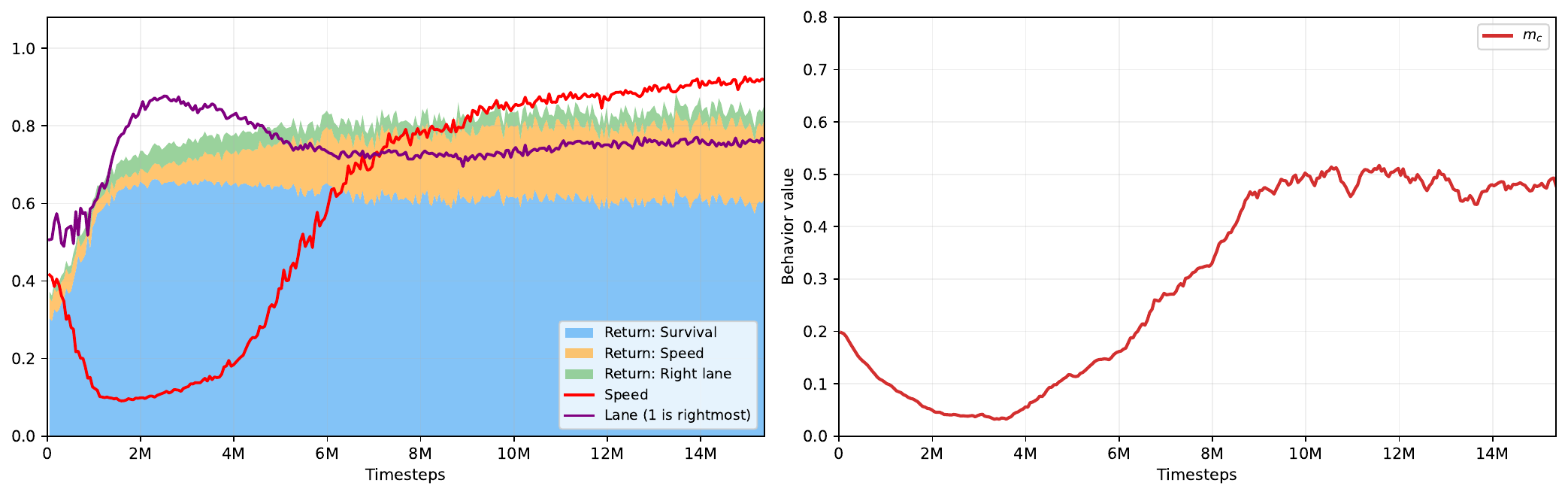}
\caption{\textbf{Left:} Return breakdown of a PPO agent training on HighJax. Training is deliberately slowed down to $D_{\mathrm{KL}} = 5 \times 10^{-4}$ per epoch. \textbf{Right:} Value of the collision behavior measure $m_c$ across training.}
\label{fig:first-run}
\end{figure}

We show an example training run on HighJax in Fig.~\ref{fig:first-run} and offer a simplified analysis. The dominant reward component is collision, and the agent first optimizes for that. In the first 1M timesteps, the agent learns two simple ways to avoid collision: slow down and stick to the rightmost lane. Survival climbs from 0.04 to 0.37. Between 1M and 4M timesteps, the agent learns to stay in the right lane more often, increasing survival to 0.85. From 4M timesteps onward, the agent learns to speed up. This leads to more collisions with other vehicles, but these are few enough that the penalties are offset by the reward from the increased speed.

In this simple case, the agent's change towards more collisions is easily explained: the return that it gets from the extra speed is greater than the loss from the collision. However, let us practice BXRL by defining a behavior measure $m_c$ to capture this behavior. We cherry-pick six observations in which the agent chose actions \textsc{left}, \textsc{right}, or \textsc{faster}, each of which led to a collision two timesteps later. We note that this is one of many overlapping ways to express this behavior. Formally:
\begin{equation}
m_c(\pi) \;=\; \frac{1}{3}\!\left(\,\mathbb{E}_{o \sim \mathcal{D}_c^L}\!\left[\pi_\theta(\textsc{left} \mid o)\right] \;+\; \mathbb{E}_{o \sim \mathcal{D}_c^R}\!\left[\pi_\theta(\textsc{right} \mid o)\right] \;+\; \mathbb{E}_{o \sim \mathcal{D}_c^F}\!\left[\pi_\theta(\textsc{faster} \mid o)\right]\right)
\label{eq:m-collision}
\end{equation}
where $\mathcal{D}_c^L$, $\mathcal{D}_c^R$, and $\mathcal{D}_c^F$ are each uniform over the two scenarios for their respective action. We show the example scenarios in Fig.~\ref{fig:copper-scenarios}, their full observation matrices in Appendix~\ref{app:observations}, and the behavior's value throughout the training in Fig.~\ref{fig:first-run} (right).

\begin{figure}[t]
\centering
\includegraphics[width=\textwidth]{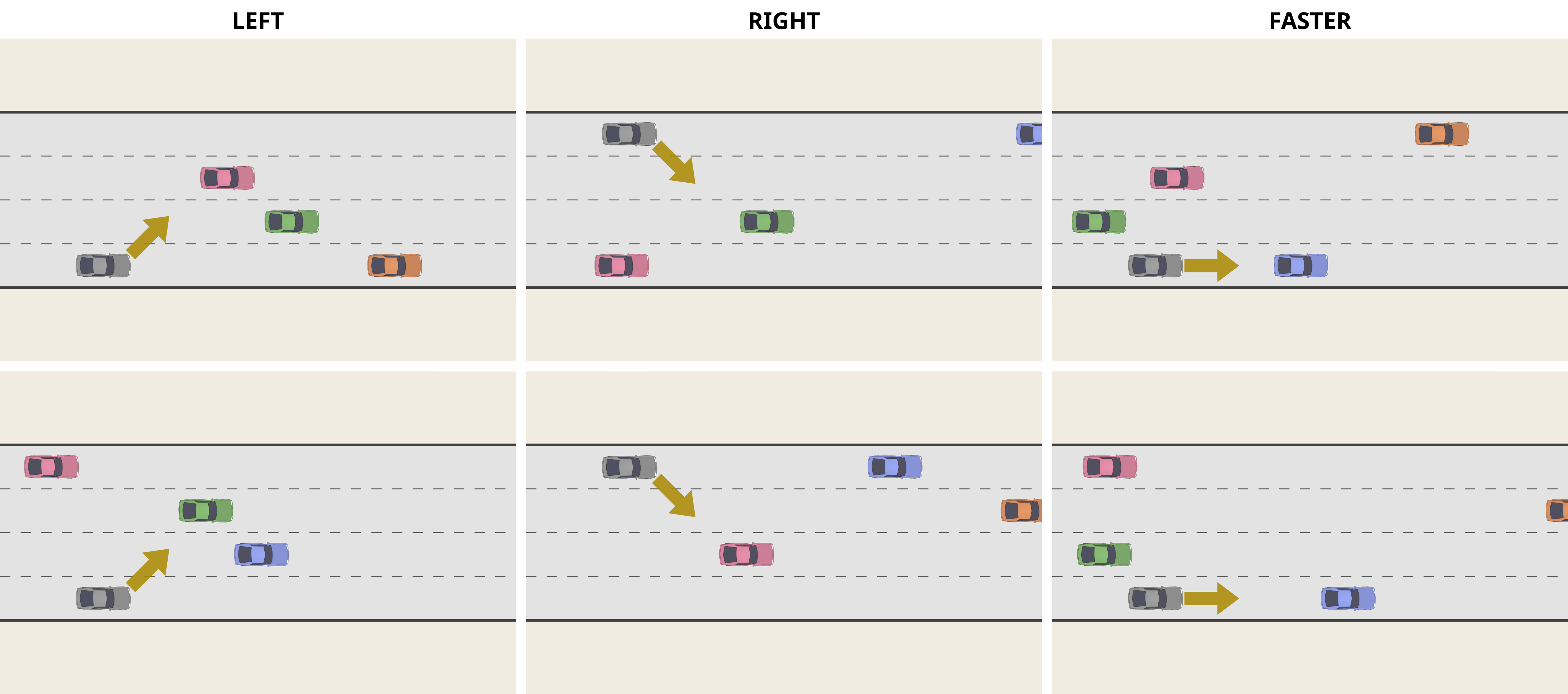}
\caption{Six scenarios defining the collision behavior measure $m_c$. Each causes the ego car (dark gray) to crash within two timesteps. Columns correspond to the action taken: \textsc{left}, \textsc{right}, or \textsc{faster}.}
\label{fig:copper-scenarios}
\end{figure}

\textbf{We present this collision behavior, as measured by $m_c$, as an example target for potential BXRL methods.} Our goal here is to provide only a proof of concept measure that could be explained with one of the methods we describe in Section~\ref{sec:adapting-methods}. We leave the implementation of such adaptations and the analysis of their subsequent explanations to future work.

\section{Limitations}
\label{sec:limitations}

BXRL requires practitioners to choose which observations to sample for a given behavior measure. This can be very difficult, especially when the observation space is large. We consider the example behavior of overtaking another vehicle, and highlight three challenges.

\begin{enumerate}
    \item There could be many different trajectories for accomplishing what humans perceive as the same behavior. In the overtaking example, the ego vehicle is behind another vehicle, and then it changes lanes to the left. Consider that one policy might choose to execute the left turn when the other vehicle is 20 meters ahead, while another policy will prefer 10 meters. Both policies are considered to overtake, but if a behavior measure is defined in terms of the former policy, e.g., $m_o(\pi) := \pi(a_\textsc{left} \mid o_\text{20m})$, it might return a misleadingly low value for the latter policy.
    \item A more extreme case of the former problem: different policies can have wildly different probability distributions for the observations that they visit. After repeated training, a policy might develop arbitrary action distributions for observations it never visits; these action distributions might choose nonsensical behavior. When defining a behavior, it may be challenging to choose observations that do not belong to the unvisited set of any of the hundreds of policies for which one wants to measure the behavior.
    \item Like many behaviors, overtaking requires multiple actions at different times: changing lanes to the left, then speeding up, and then changing lanes back to the right. (Putting aside cases where the overtaking fails and the vehicle needs to gracefully abort the attempt.) A good behavior measure will need to combine the action distributions for each of these observations.
\end{enumerate}

We believe that these problems are solvable, though they would require behavior measures that are far more involved than the basic ones we demonstrated in Section~\ref{sec:express-behavior}. We outline potential solutions to these three challenges, drawing on adjacent literatures. For Challenge~1, the quality-diversity literature faces the same problem when defining behavior descriptors across diverse policies; \citet{wu2023quality} derive gradient estimators that make behavior descriptors robust to trajectory variation, and contrastive or bisimulation-based embeddings \citep{laskin2022cic,zhang2021bisimulation} produce spaces in which behaviorally equivalent states cluster together regardless of the specific trajectory taken. For Challenge~2, the DICE family of stationary distribution correction estimators \citep{nachum2019dualdice,zhang2020gendice} computes the density ratio $d^\pi(s,a) / d^{\mathcal{D}}(s,a)$ without requiring knowledge of the data-collection policy, enabling behavior measures to be reweighted to a common reference distribution; distributionally robust evaluation \citep{shi2024robust} can further provide guaranteed bounds on $m(\pi)$ under worst-case perturbations. For Challenge~3, the options framework \citep{sutton1999between} provides a natural grammar in which primitive sub-behaviors are defined as options and composed via linear temporal logic \citep{araki2021logical}, and data-driven methods such as PRISE \citep{zheng2024prise} can discover such primitives automatically by applying byte-pair encoding to action sequences. These directions are complementary: temporal abstraction decomposes a complex behavior into measurable primitives, distribution correction enables comparison across policies, and robust observation sampling groups diverse implementations of the same behavior.

\section{Discussion and Conclusion}
\label{sec:discussion}

This is a problem formulation paper; the most important kind of future work that we propose is the development and comparative evaluation of BXRL methods. We hope to inspire our fellow researchers to develop such methods, and we are eagerly anticipating to evaluate the explanations they will produce. In the meanwhile, we propose two more directions for future work below:

\paragraph{Behavior-Explainable Artificial Intelligence} The BXRL formulation can be generalized from RL to other fields in Machine Learning. In supervised learning, the policy $\pi_\theta$ is replaced by a classifier or regressor, and the behavior measure $m$ can express any pattern over its predictions. For LLMs fine-tuned with RLHF, $m$ can capture sycophancy \citep{sharma2023sycophancy}, verbosity, or refusal rates, which are quantities that practitioners already monitor. Formalizing these as behavior measures and applying the BXRL framework would unify what the mechanistic interpretability community is already doing under an explicit explanatory lens. We refer to this broader program as Behavior-Explainable Artificial Intelligence (BXAI).

\paragraph{User studies} The XRL literature has a well-documented evaluation gap: \citet{milani2024explainable} found that comprehensibility and preferability are the least commonly evaluated metrics in XRL papers. Whether behavior-level explanations actually improve human understanding and decision-making in safety-critical domains remains an open empirical question. We echo \citet{gyevnar2025objective} in calling for objective success metrics and rigorous user studies as a prerequisite for claims of explanatory utility.

To conclude, we argue that \textit{behavior} is a missing level of abstraction in explainable reinforcement learning. Situated between single-action explanations and global policy summaries, behaviors capture the recurring patterns that characterize what an agent tends to do. By defining the properties of behaviors and showing how existing XRL methods can be adapted to reveal them, this paper lays the groundwork for BXRL and invites the community to develop and rigorously evaluate behavioral explanations.


\makeatletter
\if@accepted
\section*{Acknowledgments}

We thank our colleagues for helpful discussions and feedback: David Aha, Nitay Alon, Ofra Amir, Eli Bronstein, Bálint Gyevnár, Alexandre Heuillet, David Manheim, Stephanie Milani, Tim Miller, and Georg Ostrovski.
\fi
\makeatother


\bibliography{bibliography}
\bibliographystyle{rlj}

\appendix

\section{HighJax Environment Rules}
\label{app:env-details}

HighJax is a port of the \texttt{highway-v0} setting, in which the agent controls a vehicle navigating a four-lane highway with fifty other vehicles ahead. The agent's vehicle is also called the \textit{ego vehicle}.

\paragraph{Action space} The \texttt{highway-v0} setting has a discrete action space: \textsc{left}, \textsc{idle}, \textsc{right}, \textsc{faster} and \textsc{slower}. These are interpreted by the environment into smooth movements. The environment state includes a discrete \textit{target lane} and \textit{target speed}. The latter is one of three options: 20 m/s, 25 m/s, and 30 m/s. Each action increments, decrements or sustains either the target lane or speed. The target lane and speed are interpreted by two proportional controllers into smooth acceleration and steering movements. These controllers operate 15 times per timestep.

\paragraph{Other vehicles} The other vehicles are controlled by three predefined algorithms: (1) acceleration is determined by the Intelligent Driver Model (IDM) algorithm \citep{treiber2000idm}, (2) high-level lane-changing decisions are determined by the MOBIL algorithm \citep{kesting2007mobil}, and (3) low-level steering actions are determined by a proportional controller. Each vehicle has randomized IDM parameters for variety.

\paragraph{Reward structure} The agent is rewarded for avoiding collisions, driving fast, and staying as close to the rightmost lane as possible. The reward at each timestep is
\begin{equation*}
    r_t = -1 \cdot \mathds{1}[\text{collision}] + 0.4 \cdot \tilde{v}_t + 0.1 \cdot \tilde{\ell}_t
\end{equation*}
where $\tilde{v}_t = \text{clip}((v_t - 20)/10,\; 0,\; 1)$ is the normalized forward speed (0 at $\leq$20\,m/s, 1 at $\geq$30\,m/s) and $\tilde{\ell}_t = \ell_t / 3$ is the normalized lane index (0 for leftmost, 1 for rightmost). For benchmarking we report the normalized reward $(r_t + 1)/1.5 \in [0, 1]$.

\paragraph{Collision detection} Collisions are detected via the Separating Axis Theorem on oriented bounding boxes, and they terminate the episode.

\paragraph{Observations} The observation is a matrix $\mathbf{O} \in \mathbb{R}^{5 \times 5}$ with the following data for the ego vehicle and the four closest vehicles ahead: presence, $x$, $y$, $v_x$, and $v_y$. The rows for non-ego vehicles are in relative coordinates.

\section{Collision Behavior Observations}
\label{app:observations}

Table~\ref{tab:copper2-observations} lists the six observation matrices used to define the collision behavior measure $m_c$ (Eq.~\ref{eq:m-collision}). Each observation was selected from a training run at the indicated epoch and timestep, corresponding to a moment two timesteps before a collision. All values are normalized: positions and velocities for non-ego vehicles are relative to the ego vehicle.

\begin{table}[h]
\caption{Observation matrices for the six scenarios defining the collision behavior measure $m_c$ (Eq.~\ref{eq:m-collision}). Each matrix $\mathbf{O} \in \mathbb{R}^{5 \times 5}$ contains the ego vehicle (absolute) and four nearest NPCs (relative). Columns: presence, longitudinal position, lateral position, longitudinal velocity, lateral velocity.}
\label{tab:copper2-observations}
\centering
\begin{subtable}[t]{0.48\textwidth}
\centering
\caption{Scenario 1: \textsc{left} (epoch 148, $t$\,=\,226)}
\scriptsize
\begin{tabular}{l rrrrr}
\toprule
 & pres. & $x$ & $y$ & $v_x$ & $v_y$ \\
\midrule
ego & 1.000 & 1.000 & 0.750 & 0.373 & 0 \\
npc$_0$ & 1.000 & 0.057 & -0.500 & -0.089 & 0 \\
npc$_1$ & 1.000 & 0.086 & -0.250 & -0.091 & 0 \\
npc$_2$ & 1.000 & 0.133 & 0 & -0.074 & 0 \\
npc$_3$ & 1.000 & 0.636 & -0.500 & -0.088 & 0 \\
\bottomrule
\end{tabular}
\end{subtable}
\hfill
\begin{subtable}[t]{0.48\textwidth}
\centering
\caption{Scenario 2: \textsc{left} (epoch 221, $t$\,=\,118)}
\scriptsize
\begin{tabular}{l rrrrr}
\toprule
 & pres. & $x$ & $y$ & $v_x$ & $v_y$ \\
\midrule
ego & 1.000 & 1.000 & 0.750 & 0.375 & 0 \\
npc$_0$ & 1.000 & -0.024 & -0.750 & -0.107 & 0 \\
npc$_1$ & 1.000 & 0.047 & -0.500 & -0.113 & 0 \\
npc$_2$ & 1.000 & 0.072 & -0.250 & -0.102 & 0 \\
npc$_3$ & 1.000 & 0.336 & -0.750 & -0.106 & 0 \\
\bottomrule
\end{tabular}
\end{subtable}

\vspace{1em}

\begin{subtable}[t]{0.48\textwidth}
\centering
\caption{Scenario 3: \textsc{right} (epoch 92, $t$\,=\,102)}
\scriptsize
\begin{tabular}{l rrrrr}
\toprule
 & pres. & $x$ & $y$ & $v_x$ & $v_y$ \\
\midrule
ego & 1.000 & 1.000 & 0 & 0.311 & 0 \\
npc$_0$ & 1.000 & -0.003 & 0.750 & -0.036 & 0 \\
npc$_1$ & 1.000 & 0.063 & 0.500 & -0.061 & 0 \\
npc$_2$ & 1.000 & 0.189 & 0 & -0.057 & 0 \\
npc$_3$ & 1.000 & 0.326 & 0.250 & -0.048 & 0 \\
\bottomrule
\end{tabular}
\end{subtable}
\hfill
\begin{subtable}[t]{0.48\textwidth}
\centering
\caption{Scenario 4: \textsc{right} (epoch 111, $t$\,=\,142)}
\scriptsize
\begin{tabular}{l rrrrr}
\toprule
 & pres. & $x$ & $y$ & $v_x$ & $v_y$ \\
\midrule
ego & 1.000 & 1.000 & 0.003 & 0.323 & -0.002 \\
npc$_0$ & 1.000 & 0.053 & 0.497 & -0.063 & 0.002 \\
npc$_1$ & 1.000 & 0.121 & -0.003 & -0.066 & 0.002 \\
npc$_2$ & 1.000 & 0.182 & 0.247 & -0.055 & 0.002 \\
npc$_3$ & 1.000 & 0.335 & 0.747 & -0.073 & 0.002 \\
\bottomrule
\end{tabular}
\end{subtable}

\vspace{1em}

\begin{subtable}[t]{0.48\textwidth}
\centering
\caption{Scenario 5: \textsc{faster} (epoch 81, $t$\,=\,233)}
\scriptsize
\begin{tabular}{l rrrrr}
\toprule
 & pres. & $x$ & $y$ & $v_x$ & $v_y$ \\
\midrule
ego & 1.000 & 1.000 & 0.750 & 0.259 & 0 \\
npc$_0$ & 1.000 & 0.010 & -0.500 & 0.001 & 0 \\
npc$_1$ & 1.000 & -0.026 & -0.250 & 0.013 & 0 \\
npc$_2$ & 1.000 & 0.066 & 0 & 0.005 & 0 \\
npc$_3$ & 1.000 & 0.131 & -0.750 & -0.003 & 0 \\
\bottomrule
\end{tabular}
\end{subtable}
\hfill
\begin{subtable}[t]{0.48\textwidth}
\centering
\caption{Scenario 6: \textsc{faster} (epoch 117, $t$\,=\,130)}
\scriptsize
\begin{tabular}{l rrrrr}
\toprule
 & pres. & $x$ & $y$ & $v_x$ & $v_y$ \\
\midrule
ego & 1.000 & 1.000 & 0.750 & 0.321 & 0 \\
npc$_0$ & 1.000 & -0.021 & -0.750 & -0.066 & 0 \\
npc$_1$ & 1.000 & -0.023 & -0.250 & -0.067 & 0 \\
npc$_2$ & 1.000 & 0.088 & 0 & -0.066 & 0 \\
npc$_3$ & 1.000 & 0.191 & -0.500 & -0.053 & 0 \\
\bottomrule
\end{tabular}
\end{subtable}
\end{table}

\end{document}